\pdfoutput=1

\documentclass[11pt]{article}

\usepackage[final]{acl}

\usepackage{times}
\usepackage{latexsym}

\usepackage[T1]{fontenc}

\usepackage[utf8]{inputenc}

\usepackage{microtype}

\usepackage{inconsolata}
\usepackage{graphicx}
\usepackage{caption}
\usepackage{stmaryrd}
\usepackage{amsmath}
\usepackage{booktabs}

\title{Self-Regulated Data-Free Knowledge Amalgamation for Text Classification}

\author{Prashanth Vijayaraghavan \\
  IBM Research \\
  San Jose CA 95120 \\
  \texttt{prashanthv@ibm.com} \\\And
  Hongzhi Wang \\
  IBM Research \\
  San Jose CA 95120 \\
  \texttt{hongzhiw@us.ibm.com} \\\And
  Luyao Shi \\
  IBM Research \\
  San Jose CA 95120 \\
  \texttt{luyao.shi@ibm.com} \\
  \AND
  Tyler Baldwin \\
  IBM Research \\
  San Jose CA 95120 \\
  \texttt{tbaldwin@us.ibm.com} \\\And
  David Beymer \\
  IBM Research \\
  San Jose CA 95120 \\
  \texttt{beymer@us.ibm.com} \\\And
  Ehsan Degan \\
  IBM Research \\
  San Jose CA 95120 \\
  \texttt{edehgha@us.ibm.com} \\}

\begin{document}
\maketitle

\begin{abstract}
Recently, there has been a growing availability of pre-trained text models on various model repositories. These models greatly reduce the cost of training new models from scratch as they can be fine-tuned for specific tasks or trained on large datasets. However, these datasets may not be publicly accessible due to the privacy, security, or intellectual property issues. In this paper, we aim to develop a lightweight student network that can learn from multiple teacher models without accessing their original training data. Hence, we investigate Data-Free Knowledge Amalgamation (\textsc{Dfka}), a knowledge-transfer task that combines insights from multiple pre-trained teacher models and transfers them effectively to a compact student network. To accomplish this, we propose \textsc{StrataNet}, a modeling framework comprising: (a) a steerable data generator that produces text data tailored to each teacher and (b) an amalgamation module that implements a self-regulative strategy using confidence estimates from the teachers' different layers to selectively integrate their knowledge and train a versatile student. We evaluate our method on three benchmark text classification datasets with varying labels or domains. Empirically, we demonstrate that the student model learned using our \textsc{StrataNet} outperforms several baselines significantly under data-driven and data-free constraints.
\end{abstract}

\section{Introduction}
Recent NLP advancements have yielded numerous pre-trained models, often achieving state-of-the-art performance across various tasks. These models are publicly available to promote reproducibility and further research. To facilitate knowledge transfer from pre-trained teacher models, \citet{hinton2015distilling} pioneered Knowledge Distillation (KD), utilizing soft target labels to train light-weight student models effectively. Subsequently, diverse KD approaches have been successfully applied in different domains. Traditionally, KD relies on using original training data to guide the student model's learning from a task-specific teacher model. However, this approach has limitations, often involving learning from a single teacher model \cite{sanh2019distilbert,liu-etal-2020-fastbert} or a task-specific ensemble of teachers \cite{fukuda2017efficient,tian2019contrastive}.


\begin{figure}[t!]
\centering
\includegraphics[width=0.5\textwidth]{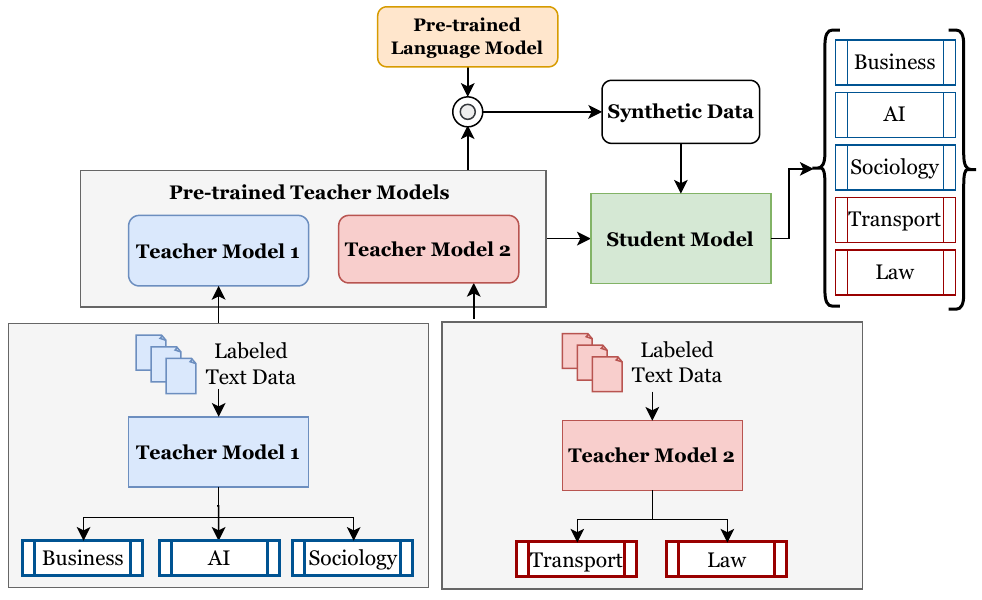}
\caption{Given a set of pre-trained teacher models (Teacher Models 1 \& 2), each with distinct expertise, the goal is to train a student model capable of amalgamating their knowledge, mastering prediction across all specialized classes of the teachers. \label{fig:dfka_ov}}
\end{figure}

Unlike traditional KD, where teachers focus on the same task, knowledge amalgamation (KA) techniques \cite{luo2019knowledge,shen2019amalgamating} enable learning in a student network by integrating knowledge from multiple teachers with diverse expertise. These methods enhance the student model's classification abilities across a wider range of labels. While KA techniques are well-established in Computer Vision, their exploration in NLP literature is limited. \citet{li-etal-2022-mimicking} utilized Monte-Carlo Dropout to estimate model uncertainty for merging knowledge from different pre-trained teacher models. However, these techniques often require access to unlabeled data from the original training set used by the pre-trained models \cite{luo2019knowledge,shen2019amalgamating,li2021model,vongkulbhisal2019unifying} to train a versatile student model. Unfortunately, the original training data and annotations are often unavailable due to various issues. Moreover, the diverse expertise of teacher models may lead to uncertain states and probabilities when handling input sequences outside their domains. These challenges hinder the application of KA methods in broader domains. To address this, we explore a practical knowledge-transfer task called Data-Free Knowledge Amalgamation (\textsc{Dfka}). Figure \ref{fig:dfka_ov} provides an overview of this task, aiming to enhance the student model's capabilities by integrating knowledge from multiple pre-trained teachers without access to the original training data.

To achieve our goal, we introduce \textsc{StrataNet}\footnote{Short for \textbf{S}elective \textbf{T}ransformer based Self-\textbf{R}egul\textbf{AT}ive \textbf{A}malgamation \textbf{NET}work}, a knowledge amalgamation framework with: (i) a flexible generation module creating pseudo text data for each pre-trained teacher network, and (ii) an amalgamation module enabling self-regulated integration of teachers' knowledge during student model training. Integration is guided by a teacher-specific out-of-distribution (OOD) score, assessing the reliability of intermediate and output states of every pre-trained teacher model. \\\textbf{Contributions}:
(1) Introduction of \textsc{StrataNet}, a pioneering data-free knowledge amalgamation (\textsc{DFKA}) method for lightweight student model training without accessing original training data.
(2) Proposal of a block-wise amalgamation strategy for integrating knowledge from multiple heterogeneous (or homogeneous) teacher model layers into the student model.
(3) Demonstration of superior performance by our \textsc{StrataNet}-trained student model compared to various baselines across three benchmark text datasets: AG News, OhSumed Abstracts, and 5 Abstracts Group.

\section{Related Work}
In this section, we explore the relevant literature concerning knowledge distillation (KD) and amalgamation. KD is a technique aimed at transferring knowledge from a large teacher network to a student model, offering benefits across various NLP tasks and facilitating model compression. These tasks encompass question answering \cite{izacard2020distilling,yang2020model}, multi-modal summarization \cite{zhang2022unims}, and neural machine translation \cite{tan2019multilingual,wang2021selective,zhou2019understanding}, among others. Notable approaches such as DistilBERT \cite{sanh2019distilbert} and TinyBERT \cite{jiao2019tinybert} primarily focus on compressing models, maintaining the student architecture identical to that of the teacher model (i.e., homogeneous setting). Fewer models, like those by Tang et al. \cite{tang2019distilling,tang2019natural}, train a heterogeneous student model. While KD has found widespread application in NLP, data-free knowledge distillation (DFKD) remains relatively underexplored compared to its application in computer vision. Recent studies \cite{melas2020generation,ma2020adversarial,ma2022prompting} have delved into training compressed student models under data-free settings using techniques such as training data augmentation, plug \& play embedding guessing, and reinforced topic prompter.

In contrast to the singular teacher model approach in KD, knowledge amalgamation (KA) involves training a versatile student model by amalgamating insights from multiple pre-trained teacher models. \citet{li-etal-2022-mimicking} utilized Monte Carlo Dropout to estimate model uncertainty and perform classification on the union of label sets from different teacher models. Although these methods do not rely on human-annotated labels, they leverage input text from the original training data. \citet{jin2022dataless} proposed a parameter space merging method for dataless knowledge fusion, assuming an impractical uniformity in model architectures across input and merged models. Differing from the aforementioned approaches, our method, StrataNet, introduces a framework for data-free knowledge amalgamation (DFKA) in text, representing a pioneering exploration in NLP literature involving multiple heterogeneous teacher networks.

\section{Problem Setup}
\label{sec:prob_st}
Given $K$ pre-trained teacher models $\mathcal{T} = \{\mathcal{T}_i\}_{i=1}^K$, each with $L_{\mathcal{T}_i}$-layers and its own domain of expertise, i.e., performing a $c_i$-class classification task with few overlapping or disjoint set of labels $\mathcal{Y}_i =\{y_i^j\}_{j=1}^{c_i}$, our goal is to train a lightweight student model $\mathcal{S}$ with $L_{\mathcal{S}}$-layers such that it can compute predictions over the union of all the label sets, $\mathcal{Y}=\bigcup_{i=1}^{K}{\mathcal{Y}_i}$ and $L_{\mathcal{S}} \leq \min(\{L_{\mathcal{T}_i}\}_{i=1}^K$).


\section{Proposed Approach}
\begin{figure*}[t]
\includegraphics[width=0.9\textwidth]{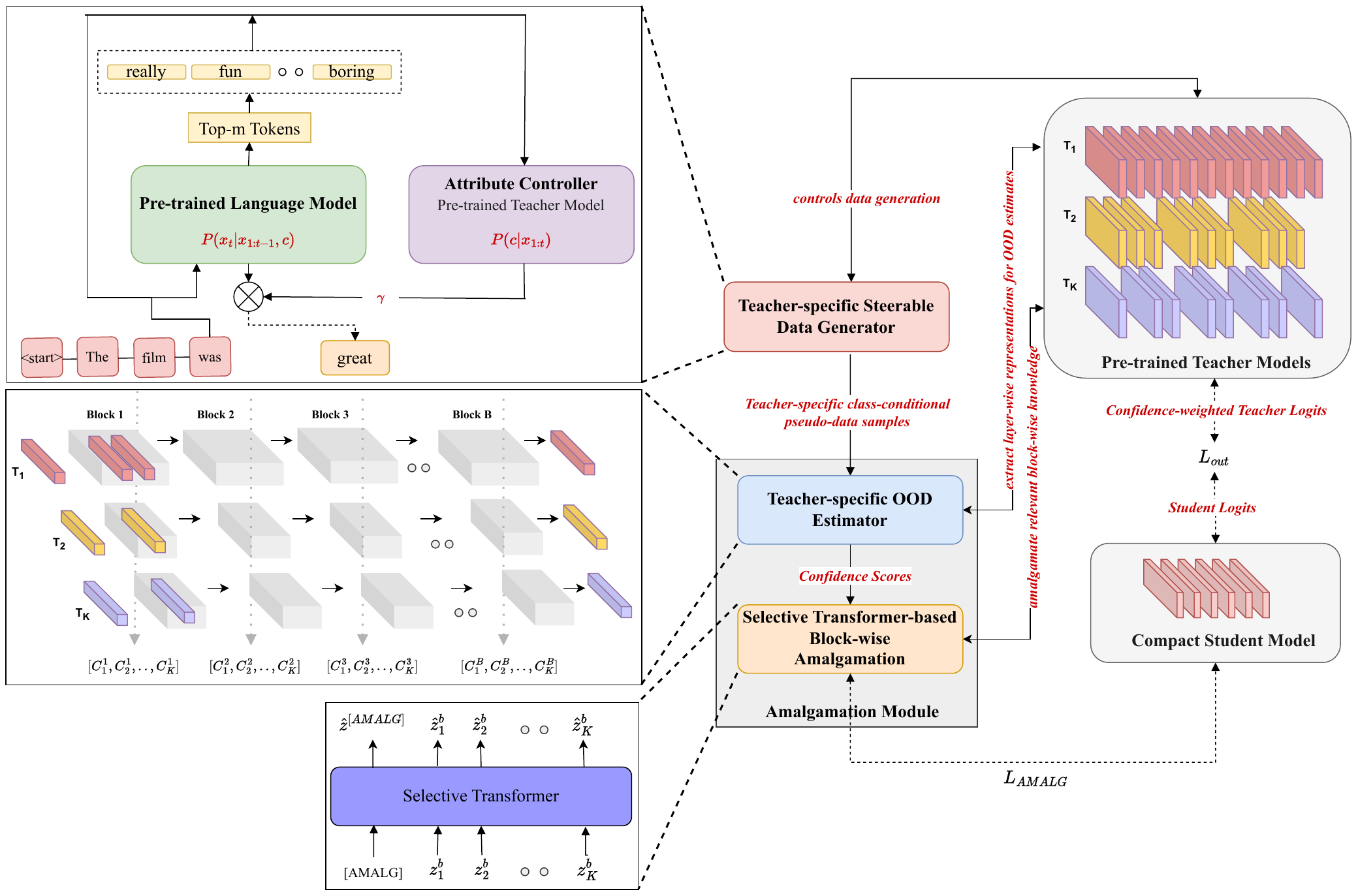}
\caption{Illustration of our \textsc{StrataNet} framework. \label{fig:illus}}
\end{figure*}
\subsection{Overview}
In this section, we outline our framework, \textsc{StrataNet}, designed to train a lightweight student model using multiple teachers under data-free constraints. We address the following factors: (a) lack of training data, (b) existence of specialized teachers with non-overlapping or partially overlapping label sets, and (c) need to integrate knowledge from diverse teachers. Our \textsc{StrataNet} consists of two main components. The first, $\mathcal{G}_i$, is a teacher-specific steerable data generator. It guides a base pre-trained language model, $\mathcal{P}$, to generate tailored text for each teacher, $\mathcal{T}_i$, overcoming data scarcity by creating pseudo-data samples. The second component, the amalgamation module, serves two functions. It evaluates each teacher's confidence in predicting within their expertise and employs block-wise integration with a selective transformer to fuse knowledge from multiple teachers. Utilizing confidence scores, this approach appropriately weights representations from different teacher models, effectively managing diverse teacher architectures.

\subsection{Steerable Data Generator}


 To overcome the challenge of unavailability of the original training data for teacher models, we utilize a conditional text generation method that generates pseudo-data samples specifically tailored to the label set of the teacher $\mathcal{T}_i$. Given a teacher model $\mathcal{T}_i$ and any class label $c \in \mathcal{Y}_i$, a steerable text generator, $\mathcal{G}_i$, produces a class-controlled text $x$ of length $N$ as follows: $P(x_{1:N}|c) = \prod_{t=1}^{N}{P(x_t|x_{1:t-1}, c)}$


For each teacher $\mathcal{T}_i$, our steerable text generator produces pseudo-data samples $\mathcal{D}_i^p = (\mathcal{X}_i^p, \hat{\mathcal{Y}}_i^p)$ by applying an inference-time controllable generation method to steer an unconditional language model towards the desired class label relevant to a specific teacher. The generation process entails guiding a base pre-trained language model (PLM), denoted as $\mathcal{P}$, using a post-processing module. By adjusting the parameters during the decoding phase, the generator exhibits varying degrees of class control over the text sampled from the chosen base PLM.



Based on a recent study by \citet{gu2022improving}, we adopt a variant of the weighted decoding method to generate class-conditional text using a pre-trained unconditional language model, denoted as $\mathcal{P}$. In this approach, we model the generation process by incorporating a Bayesian factorization as follows:
\begin{equation}
P(x_t|x_{1:t-1}, c) \propto P(x_t|x_{1:t-1}) P(c|x_{1:t})^\gamma
\label{eq:bayesfactor}
\end{equation}

Here, $\gamma$ represents a hyperparameter for control strength. The first term corresponds to the output probabilities generated by the chosen PLM, while the second term relies on the teacher model to estimate the likelihood of the generated text (up to the current time step $t$) being classified under the class label $c$. During the sampling process, the value of $\gamma$ regulates the influence of the teacher model. 

One challenge in this approach is the computational complexity of teacher-guided sequence sampling. To compute the second term in Equation \ref{eq:bayesfactor}, we need to estimate the class probability $P(c|x_{1:t})$, requiring evaluation of $P(c|x_{1:t-1}, x_t)$ for every token in the vocabulary $\mathcal{V}$ at the $t^{th}$ timestep. To reduce inference time, we exclude low-probability tokens and prioritize a subset for teacher guidance. Tokens with low probability $P(x_t|x_{1:t-1})$ from the PLM are discarded, even if the teacher model assigns high weights $P(c|x_{1:t})$. Consequently, we exclusively use the top-m tokens with higher probabilities, guided by the teacher model’s weights. Subsequently, we employ a top-k sampling strategy, where $k < m \ll |\mathcal{V}|$). Our experiments indicate that setting $m = 100$ is notably effective. Table \ref{tab:samplegen} displays sample generations produced using the teacher-guided generation module. In this experiment, we trained two teachers on the AG News and OhSumed label sets. Subsequently, we generated pseudo-data samples for the "Sports" category under the guidance of the teacher trained on the AG News dataset and for the "Cardiovascular Diseases" category under the guidance of the teacher trained on the OhSumed dataset. Notably, we observe that the generated text is not only fluent but also relevant to the respective category of interest.
\begin{table}[ht!]
\small
\begin{tabular}{p{0.23\linewidth} | p{0.7\linewidth}}
\toprule
Dataset & Text Generated \\ \midrule
AG News & \begin{tabular}[c]{@{}p{\linewidth}}\textbf{Category: Sports}\\ In an electrifying moment that left spectators spellbound, Olympic speedster Usain Bolt once again proved that he is the fastest man alive by shattering yet another world record. As he crossed the finish line, the Jamaican sprinter scorched towards the tape while leaving all competition behind ... In front of thousands of cheering fans, Bolt completed the race faster than his previous mark set last year. It was a performance that brought tears of joy to the eyes.... \end{tabular} \\\midrule
OhSumed & \begin{tabular}[c]{@{}p{1\linewidth}}\textbf{Category: Cardiovascular Diseases}\\ The study aimed to determine the prevalence of echocardiographic aortic regurgitation among patients presenting for screening echocardiography at a single university center. Echocardiograms were performed in accordance with Echocardiography guidelines ... New findings reveals important data addressing our knowledge gap regarding Aortic Regurgitation patient prevalence. In this study, transthoracic imaging confirmed prevalence across all ages stratified by 10yr increments and between men and women. The report sheds light on the epidemiology of AR found by echocardiogram ....\end{tabular} \\ \bottomrule
\end{tabular}
\caption{Sample Generations from our Steerable Generation Module \label{tab:samplegen}}
\end{table}
\subsection{Block-wise Amalgamation Module}


We introduce an amalgamation module that estimates out-of-distribution (OOD) scores for each teacher using a Teacher-specific OOD Estimator. It integrates selective informative states from relevant teachers based on these scores using a Selective Transformer (\textsc{ST-amalg}), transferring them to a student in a blockwise manner to accommodate varying sizes of teacher models.


\subsection{Teacher-specific OOD Estimator}
\label{sec:oodest}

Due to diverse label sets $\{\mathcal{Y}_i\}_{i=1}^{K}$ in pre-trained teacher models $\{\mathcal{T}\}_{i=1}^{K}$, any input text from an unseen category for a specific teacher is considered out-of-distribution (OOD). Extracted features from that teacher's intermediate layer may not be sufficient for effective knowledge transfer to the student model. Studies indicate: (a) Transformer-based models encode transferable features in various intermediate layers \cite{liu-etal-2019-linguistic,rogers2021primer}, and (b) final layers, especially in models like BERT, are highly task-specific \cite{kovaleva-etal-2019-revealing,rogers2021primer}. Considering these, we propose layer-wise teacher-specific lightweight OOD estimators, explained below.


\subsubsection{OOD Score Computation} 

For an input text $x \in \mathcal{X}_i$ with label $y \in \hat{\mathcal{Y}}_i$, a transformer-based pre-trained teacher model $\mathcal{T}_i$ produces contextual token-level latent embeddings at each layer $l \in L_{\mathcal{T}_i}$. These are averaged into a single latent representation $h_i^l \in \mathcal{R}^{d_i}$, where $d_i$ is the dimensions of the latent representations. To compute an OOD score for any new input $x_{new}$, we use a Mahalanobis distance (\textsc{Md}) based OOD detection technique. For an in-distribution (\textsc{Id}) dataset with $c_i$-labels associated with $\mathcal{T}_i$, the \textsc{Md} technique fits $c_i$-class conditional Gaussian distributions $\mathcal{N}(\mu_y,\Sigma)$ to each of the $c_i$ \textsc{Id} classes based on training latent representations $h_i^l$. However, \citet{ren2021simple} proposed a Relative Mahalanobis distance (\textsc{Rmd}) that outperforms \textsc{Md} in OOD detection for both near and far-OOD scenarios by calculating the distance between class-conditional Gaussians and a single background Gaussian using data from all classes.
For an input $x_{new}$ with the latent representation $\hat{h}_i^l$ at layer $l$, \textsc{Rmd} is given by:
\begin{align}
    \textsc{Rmd}_y(\hat{h}_i^l) = \textsc{Md}_y(\hat{h}_i^l) - \textsc{Md}_{bg}(\hat{h}_i^l)  \label{eq:rmd} \\
    \textsc{Md}_y(\hat{h}_i^l) = (\hat{h}_i^l - \mu_y)^T\Sigma^{-1}(\hat{h}_i^l-\mu_y)  \label{eq:md} \\
    \mathcal{C}_i^l(\hat{h}_i^l) = -min_{y}\{\textsc{Rmd}_y(\hat{h}_i^l)\} \label{eq:conf}
\end{align}
where $\mathcal{C}_i^l$ refers to the confidence score of $x_{new}$ being in-domain for $\mathcal{T}_i$ based on representation at layer $l$, $\mu_y $ is a class-conditional mean vectors and $\Sigma$ is the covariance matrix, $\textsc{Md}_{bg}$ indicates Mahalanobis distance of $h_i^l$ to the background distribution fitted to the entire training data usually. The \textsc{Rmd} score acts as a contrastive measure indicating the sample's proximity to both the training and background domains. Higher scores indicate greater out-of-distribution characteristics, resulting in lower \textsc{Id} confidence scores, $\mathcal{C}_i^l$. Alternatively, in some cases, intermediate layers can be partitioned into $B$ blocks. Applying a similar procedure as described in Equations (\ref{eq:rmd})-(\ref{eq:conf}), confidence scores can be calculated for each block. Here, $\hat{h}_i^b$ represents the latent representation for block $b$, obtained by mean pooling over the layer representations within that block. We use a held-out subset of pseudo-data samples, $\hat{\mathcal{D}}_i^p$, generated for each teacher $\mathcal{T}_i$.

\subsection{Selective Transformer-based Block-wise Amalgamation \textsc{ST-amalg}} 

To transfer knowledge from diverse, larger teachers to a lightweight student model, we align intermediate representations in a block-wise manner, accommodating the varying number of layers between them. Each teacher network $\mathcal{T}_i$ may have a different number of grouped layers. We compute confidence-aware block-wise intermediate representations, $z_i^b$, using the confidence score at each block $b$ for each teacher. Inspired by the literature on multimodal analysis \cite{urooj-etal-2020-mmft,vijayaraghavan2023m,Lin_2022_ACCV}, we consider the intermediate latent vectors from $K$ teachers, denoted as $\{z_i^b\}_{i=1}^{K}$, as a token sequence fed into a Transformer layer. We introduce a learnable special token $[AMALG]$, similar to $[CLS]$, to integrate confidence-enriched representations from teachers into a final block-level amalgamated representation, denoted as $\hat{z}_{\mathcal{T}}^b$. Therefore, we refer to this layer as the Selective Transformer-based amalgamation layer (\textsc{ST-amalg}). Formally,
\begin{align}
    z_i^b = f(h_i^b)+g(\mathcal{C}_i^b) \label{eq:enrich}\\
    \hat{z}_{\mathcal{T}}^b = \textsc{ST-amalg}(\{z_i^b\}_{i=1}^{K}) \label{eq:amalg}
\end{align}
where $f,g$ are linear layers to enrich the block-level embeddings. 

\section{Training Objectives \& Details}
To amalgamate knowledge at intermediate layers, we compute L2-normalized distance between the student's projected block-level representation and the corresponding teachers' amalgamated embedding. Formally, 

\begin{equation}
\begin{aligned}
    \mathcal{L}_{\textsc{Amal}} = \sum_{b=1}^{B}{\mathcal{L}_{\textsc{Amal}}^{b}}\\
    s.t.\quad \mathcal{L}_{\textsc{Amal}}^{b}  = ||\hat{z}_{\mathcal{S}}^b - \hat{z}_{\mathcal{T}}^b||_2^2
    \label{eq:amal}
\end{aligned}
\end{equation}

For the output prediction layer, we compute the KL divergence loss based on confidence weighted combination of Teacher models and the temperature $\tau$ as: $\mathcal{L}_{out}= KL(\hat{\mathcal{T}}(x), \hat{\mathcal{S}}, \tau)$.

     

\subsection{Training details} Given steerable data generators $\{\mathcal{G}_i\}_{i=1}^K$ tied to teachers $\{\mathcal{T}_i\}_{i=1}^K$, we produce a student training transfer set, denoted as $\mathcal{D}^p$, by combining the pseudo-data samples generated for all the labels associated with each teacher. Next, we divide the intermediate layers into $B$-blocks such that the number of layers in each block may vary according to the number of layers in the teacher model. In our experiments, the teacher models (Teacher 1 and Teacher 2) are based on $\textsc{Bert}$-base-uncased \cite{devlin2018bert}, and we set $B$ to the number of intermediate layers in the compressed student model $\mathcal{S}$, i.e., $\textsc{Bert}_6$. We then compute the number of layers within each block for each teacher accordingly. A subset of pseudo-data samples generated for each teacher $\mathcal{T}_i$, represented as $\hat{\mathcal{D}}^p_i$, is used compute the layer-wise distribution statistics for OOD estimation. Finally, we use the student training transfer set $\mathcal{D}^p$ to train the student model by: (a) computing the confidence of teachers' block-wise features in predicting each input text, (b) amalgamating the confidence-enriched representations from teachers and (c) optimizing the weighted sum of intermediate ($\mathcal{L}_{\textsc{Amal}}$) and output prediction layer ($\mathcal{L}_{out}$) losses, expressed as: 
\begin{equation}
\mathcal{L} = \lambda\cdot\mathcal{L}_{\textsc{Amal}} + (1-\lambda)\cdot\mathcal{L}_{out}
\end{equation}

\section{Experiments}
\label{sec:expt}
 Our experiments address the following research questions: \textbf{(RQ1)} How does our model compare to baseline approaches for knowledge distillation in both data-driven and data-free scenarios? \textbf{(RQ2)} What is the individual impact of each component in our model on overall performance? \textbf{(RQ3)} How does our model fare when multiple heterogeneous teachers are utilized?

\begin{table}[ht!]
\small
\centering
\begin{tabular}{p{0.2\linewidth}cccc}
\toprule
\textbf{Datasets} & \textbf{\#Classes} & \textbf{\#Train} & \textbf{\#Valid} & \textbf{\#Test}  \\ \midrule
AG News           & 4                     & 108,000          & 12,000           & 7,600           \\\midrule                                         
5Abstracts Group              & 5                     & 4,770          & 530           & 1,000  \\\midrule                                      
OhSumed           & 23                    & 3,021            & 336              & 4,043                                        \\ \bottomrule
\end{tabular}
\caption{Data Statistics of benchmark text classification datasets. \label{tab:textdatasets}}
\end{table}

\subsection{Datasets}
\label{sec:app_ds}
We evaluate our approach using the following benchmark datasets:
(a) \textbf{AG News}\footnote{\url{http://groups.di.unipi.it/~gulli/AG_corpus_of_news_articles.html}} \cite{Zhang2015CharacterlevelCN}: It consists of news articles grouped into four major classes—World, Sports, Business, and Sci/Tech.
(b) \textbf{5 Abstract Group}\footnote{\url{https://github.com/qianliu0708/5AbstractsGroup}} \cite{liu2017leveraging}: This dataset contains academic paper abstracts from five different domains—business, AI, sociology, transport, and law.
(c) \textbf{Ohsumed}\footnote{\url{https://disi.unitn.it/moschitti/corpora.htm}} \cite{joachims1998text}: It comprises medical abstracts specifically related to cardiovascular diseases. We focus on single-label text categorization and exclude documents that belong to multiple categories. The data statistics for these benchmark datasets are presented in Table \ref{tab:textdatasets}.


\begin{table}[]
\small
\begin{tabular}{cccc}
\hline
\textbf{Models} & \textbf{\begin{tabular}[c]{@{}c@{}}AG \\ News\end{tabular}} & \textbf{\begin{tabular}[c]{@{}c@{}}5Abstracts\\ Group\end{tabular}} & \textbf{OhSumed} \\ \hline
Supervised & 94.6 & 90.7 & 70.5 \\ \hline
\multicolumn{4}{c}{\textbf{Data-Driven Methods}} \\ \hline
Teacher 1* & 49.9 & 42.0 & 36.2 \\
Teacher 2* & 47.5 & 51.5 & 38.18 \\
Ensemble* & 59.8 & 62.3 & 45.48 \\
MUKA-Hard* & \begin{tabular}[c]{@{}c@{}}87.0\\ $(\pm 0.40)$\end{tabular} & \begin{tabular}[c]{@{}c@{}}79.0 \\ $(\pm 0.82)$\end{tabular} & \_\_ \\
MUKA-Soft* & \begin{tabular}[c]{@{}c@{}}87.1 \\ $(\pm 0.19)$\end{tabular} & \begin{tabular}[c]{@{}c@{}}79.3 \\ $(\pm 0.85)$\end{tabular} & \_\_ \\ \hline
\multicolumn{4}{c}{\textbf{Data-Free Methods}} \\ \hline
Teacher 1 & 45.8 & 41.75 & 32.8 \\
Teacher 2 & 46.9 & 46.88 & 35.6 \\
Ensemble & 55.86 & 53.67 & 41.94 \\ \hline
\begin{tabular}[c]{@{}c@{}}Vanilla KA \\ (R)\end{tabular} & \begin{tabular}[c]{@{}c@{}}58.9 \\ $(\pm 3.19)$\end{tabular} & \begin{tabular}[c]{@{}c@{}}56.27\\ $(\pm 2.76)$\end{tabular} & \begin{tabular}[c]{@{}c@{}}47.33 \\ $(\pm 4.41)$\end{tabular} \\ \hline
\begin{tabular}[c]{@{}c@{}}Vanilla KA \\ (CD)\end{tabular} & \begin{tabular}[c]{@{}c@{}}62.43 \\ $(\pm 2.62)$\end{tabular} & \begin{tabular}[c]{@{}c@{}}61.55 \\ $(\pm 0.91)$\end{tabular} & \begin{tabular}[c]{@{}c@{}}50.91 \\ $(\pm 2.8)$\end{tabular} \\ \hline
\textsc{As-Dfd} & \begin{tabular}[c]{@{}c@{}}74.89 \\ $(\pm 0.89)$\end{tabular} & \begin{tabular}[c]{@{}c@{}}69.83\\ $(\pm 1.06)$\end{tabular} & \begin{tabular}[c]{@{}c@{}}56.08\\ $(\pm 1.6)$\end{tabular} \\ \hline
\textbf{\begin{tabular}[c]{@{}c@{}}\textsc{Stratanet}\\ (Ours)\end{tabular}} & \textbf{\begin{tabular}[c]{@{}c@{}}88.76\\ $(\pm 0.19)$\end{tabular}} & \textbf{\begin{tabular}[c]{@{}c@{}}83.6 \\ $(\pm 0.28)$\end{tabular}} & \textbf{\begin{tabular}[c]{@{}c@{}}65.92 \\ $(\pm 0.41)$\end{tabular}} \\ \hline
\end{tabular}
\caption{Evaluation results on benchmark text classification dataset averaged over 3 runs. Our method achieve statistically significant improvements over the closest baselines ($p < 0.01$). Bold face indicates the best results and * refers to results from prior literature. \label{tab:bigres}}
\end{table}


\subsection{Baselines}
We conduct a comparative analysis of our proposed model with data-driven and data-free baselines. Here is a summary of the baselines:\\
\textbf{Teacher Models}, which are used to predict individually. We assign zero probabilities to classes outside the expertise of each teacher.
\textbf{Ensemble}, which concatenates the output logits from all the teachers to obtains predictions over all the labels $\mathcal{Y}$.
\textbf{MUKA-Hard/Soft} \cite{li2021model}, which is a data-driven KA method that uses Monte-Carlo Dropout based model uncertainty to guide the student training.
\textbf{Vanilla KA} \cite{hinton2015distilling} (R/CD): which aims to mimic the soft targets produced by the logits combination of all teacher models using KL-divergence. In a data-free scenario, we consider two settings: (i) Random Text (R): The student model is trained on text sequences constructed using randomly selected words from the vocabulary of the pre-trained teacher models; and (ii) Cross-Domain Texts (CD): The student model is trained on cross-domain text corpora like Wikitext-103.
{\textbf{\textsc{As-Dfd}}} \cite{ma2020adversarial}, which is a data-free knowledge distillation approach. We modify this model for the \textsc{Dfka} scenario by crafting pseudo-embeddings for each teacher as specified in their original study and train a student model using self-supervision and KL-divergence. 
\textbf{\textsc{StrataNet}}, which is our complete \textsc{Dfka} model that generates pseudo-data samples and leverages the produced data for knowledge amalgamation. 

\begin{figure*}[t!]
\centering
\includegraphics[width=\textwidth]{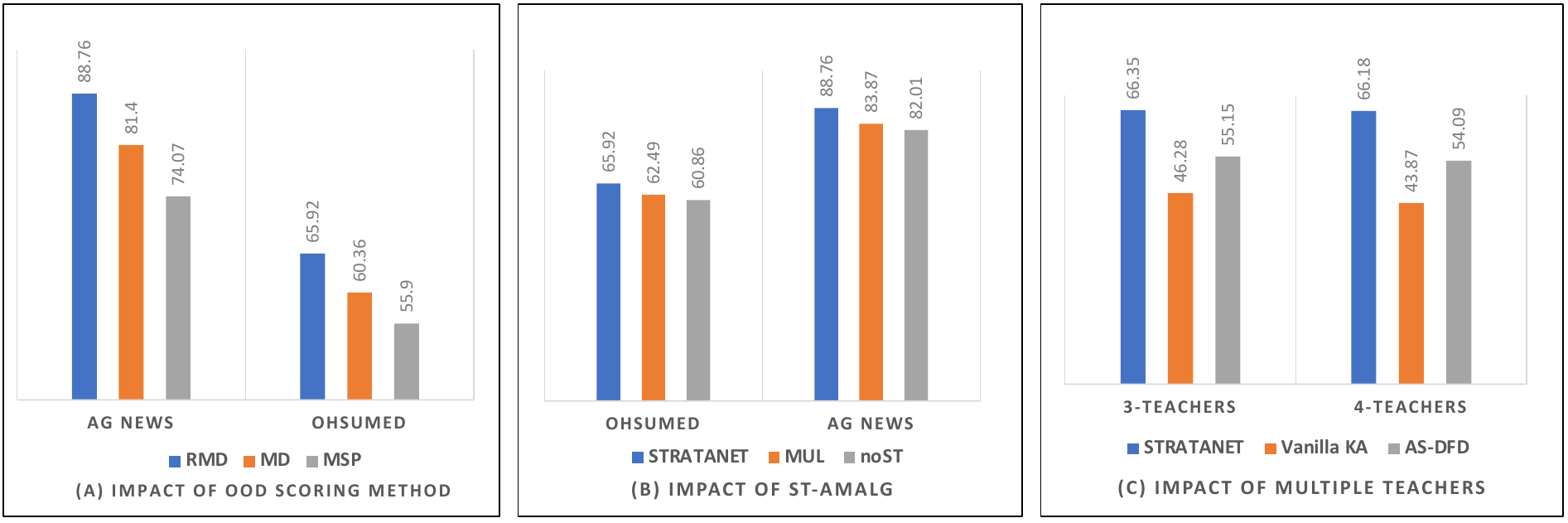}
\caption{(A) Impact of different OOD scores -- \textsc{Rmd}, \textsc{Md} \& MSP, (B) Impact of \textsc{ST-amalg}, (C) Effect of Multiple Heterogeneous teachers on OhSumed dataset.  \label{fig:charts}}
\end{figure*}

\subsection{Metrics}
To be comparable with prior studies, we compute the classification accuracy across various datasets. In particular, we report the mean and standard deviations of the accuracy over three runs in $\S$\ref{sec:res}.

\section{Results and Discussion}
\label{sec:res}

\textbf{Overall Performance} The evaluation results are presented in Table \ref{tab:bigres}, providing a summary of our findings. To ensure a fair comparison, our baselines incorporate cross-domain data (CD), similar to our model that utilizes a resource like PLM. Additionally, we implement a variation of the data-free knowledge distillation method \cite{ma2020adversarial} for \textsc{Dfka}. Compared to all the baselines, our \textsc{StrataNet} model demonstrates significant improvement over other $\textsc{Dfka}$ baselines across various text classification datasets. Notably, our compact student model trained under data-free settings shows an approximately 4\% increase in performance compared to the best-performing data-driven model in certain cases. We intuit that the knowledge from the intermediate layers are beneficial for the performance improvement.

\subsection{Ablation Studies (RQ2)}
\label{sec:ablation}


\subsubsection{Effect of \textsc{Rmd}}
In order to measure the effect of \textsc{Rmd} (explained in $\S$\ref{sec:oodest}), we replace the OOD score computation using other methods including: (a) embedding-based Mahalanobis distance (\textsc{Md}) and (b) maximum softmax probability (MSP) at the final layer. Figure \ref{fig:charts}(A) shows how modifying the OOD score has a significant impact on the overall performance of the model. RMD OOD score helps achieve the best performance of our model.


\subsubsection{Impact of \textsc{ST-amalg}}

To evaluate the contribution of \textsc{ST-amalg}, we introduce two variants: (a) \textsc{StrataNet}$_{mul}$:simply multiply the block-level confidence score with the teacher embeddings instead of the embedding enrichment (as in Equation \ref{eq:enrich}, (b) \textsc{StrataNet}$_{noST}$: remove \textsc{ST-amalg} and use a linear layer on top of confidence weighted sum of teachers' latent vectors in Equation \ref{eq:amalg}. Figure \ref{fig:charts}(B) shows that both the variants lead to significant performance degradation, asserting their value to the overall model performance. This validates our intuition that the embedding enrichment and \textsc{ST-amalg} serve as critical components to select the important block-level features from different teacher models \footnote{Additional experiments on using LLM like Llama-2 for the data generation module in Appendix \ref{sec:app_llm}}.

\subsection{Effect of Multiple Heterogeneous Teachers (RQ3)}

To demonstrate our model's ability to generalize across multiple heterogeneous teachers, we explore scenarios with three (1 BERT-base, 1 RoBerta-base, and 1 ALBERT) and four (1 BERT-base, 2 RoBerta-base, and 1 ALBERT) teachers, each with different architectures. Results are shown in Figure \ref{fig:charts}(C). While baseline KA methods struggle with increased teacher diversity, our approach consistently improves accuracy and maintains performance with more teachers. These findings underscore the robustness and effectiveness of our method across diverse experimental setups.

\section{Conclusion}
In this study, we introduce Data-Free Knowledge Amalgamation (DFKA), a method to train a lightweight student network from diverse teacher models without their original training data. Our framework, \textsc{StrataNet}, employs a steerable data generator and an amalgamation module for effective knowledge transfer. Experimental results on text datasets demonstrate the superiority of \textsc{StrataNet} over various baselines, both in data-driven and data-free scenarios. Ablation studies highlight the importance of different model components. This work opens avenues for efficient knowledge transfer in text classification, offering practical solutions for resource-constrained environments.


\section*{Limitations}
While our \textsc{StrataNet} model outperforms existing baselines, it has certain limitations. The steerable generation module, which guides text generation for specific classes, may not consistently produce accurate class-specific text. Moreover, it may not capture the full diversity of complex training datasets. Further research is needed to investigate and improve the generation module. Additionally, there is potential to expand knowledge amalgamation to tasks beyond text classification, which warrants future research.

\section*{Ethics Statement}

Our \textsc{StrataNet} model focuses on improving the performance of \textsc{Dfka} and does not introduce new ethical concerns compared to other KD/KA methods. However, we want to acknowledge two key risks here: (a) data-free knowledge amalgamation strategies can potentially be used as a precursor to model extraction attacks, compromising the confidentiality of blackbox models, as demonstrated in \cite{truong2021data}, and (b) model compression itself may introduce biases, as suggested by \cite{hooker2020characterising}. It is important to address these risks, which are not specific to our method but are common in data-free model compression techniques, in future research.

\bibliography{latex/acl_latex}

%



\clearpage
\appendix

\section{Implementation Details}
\label{sec:impl}

We base our \textsc{StrataNet} implementation on PyTorch\footnote{https://pytorch.org/}, Huggingface \cite{wolf2019huggingface} and PyTorch Lightning\footnote{https://www.pytorchlightning.ai/}. We tune our model hyperparameters using grid-search. For the generation module, we sample a maximum of 128 tokens. The top 200 tokens were selected using the nucleus sampling method with a sampling threshold of $p=0.9$. For Ohsumed dataset, we used BioGPT \cite{luo2022biogpt} in order to tailor the data generation process to the domain of interest. Trained on large-scale PubMed abstracts, BioGPT is a specialized Transformer language model designed for generating and mining biomedical text. In our experiments, we use a compressed $\textsc{Bert}$ model with 6 layers, referred to as $\textsc{Bert}_6$, as our student model. Table \ref{tab:hyperparams} shows the tuned hyperparameters used by both the generation and distillation component of our \textsc{StrataNet} model.  Our method trains a compressed student model (e.g., $\textsc{Bert}_6$) using a confidence score that selectively amalgamates the knowledge from intermediate and output layers of multiple teachers.

\begin{table}[ht!]
\small
\centering
\begin{tabular}{@{}lr@{}}
\toprule
\textbf{Hyperparameter}  & \textbf{Value}         \\ \midrule
Pre-trained LM           & GPT-2 (S/M/L) or BioGPT          \\
Learning Rate            & 2e-5                   \\
Batch Size               & 16                    \\
\#Epochs                  & 10                     \\
Dropout                  & 0.2                    \\
Optimizer                & AdamW                  \\
Learning Rate Scheduling & linear                 \\
Weight Decay             & 0.01                   \\
Warmup                   & 2 epochs               \\
Gradient Clipping        & 1.0                    \\ 
Sampling Method          & Nucleus              \\
Sampling - $p$           & 0.9                  \\
KD Temperature - $\tau$     & 0.75                \\
\bottomrule
\end{tabular}
\caption{Hyperparameters used by different components of our proposed \textsc{ProDGen} model. \label{tab:hyperparams}}
\end{table}

\section{Ablation Studies}
\begin{table}[]
\small
\centering
\begin{tabular}{@{}lcc@{}}
\toprule
\textbf{Models} & \textbf{Homogeneous} & \textbf{Heterogeneous} \\ \midrule
Teacher 1 & 49.8 & 48.9 \\
Teacher 2 & 48.86 & 50.6 \\
Ensemble & 60.25 & 60.54 \\ \midrule
\textsc{As-Dfd}$_6$ & 75.16 & 63.89 \\
\textsc{StrataNet}$_6$ &\textbf{ 89.16} & \textbf{88.53 }\\\midrule
\textsc{As-Dfd}$_4$ & 72.80 & 61.72 \\
\textsc{StrataNet}$_4$ & \textbf{88.29 }& \textbf{86.65} \\ \bottomrule
\end{tabular}
\caption{Ablation Study: Effect of heterogeneous teachers \& number of student layers \label{tab:hetlayers}}
\end{table}

\subsection{Effect of heterogeneous teachers and student model layers} In Section $\S$\ref{sec:expt}, we conducted experiments using a compressed $\textsc{Bert}_6$ model, and the results demonstrated no significant performance degradation. To delve deeper, we run additional experiments involving $\{6,4\}$-layer student models with different teacher configurations: a homogeneous setting ($\mathcal{T}_1, \mathcal{T}_2$: $\textsc{Bert}_{large}$) and a heterogeneous setting ($\mathcal{T}_1$: $\textsc{Bert}_{large}$, $\mathcal{T}_2$: $\textsc{RoBerta}_{large}$). The evaluations on the AG News dataset reveal the poor performance of the data-free baseline $\textsc{As-Dfd}$ with compressed layers, highlighting the challenges of the heterogeneous setting. However, our  \textsc{StrataNet} framework demonstrates consistent and robust performance under both configurations, even with higher compression. 

\begin{figure}[t!]
\centering
\includegraphics[width=0.4\textwidth]{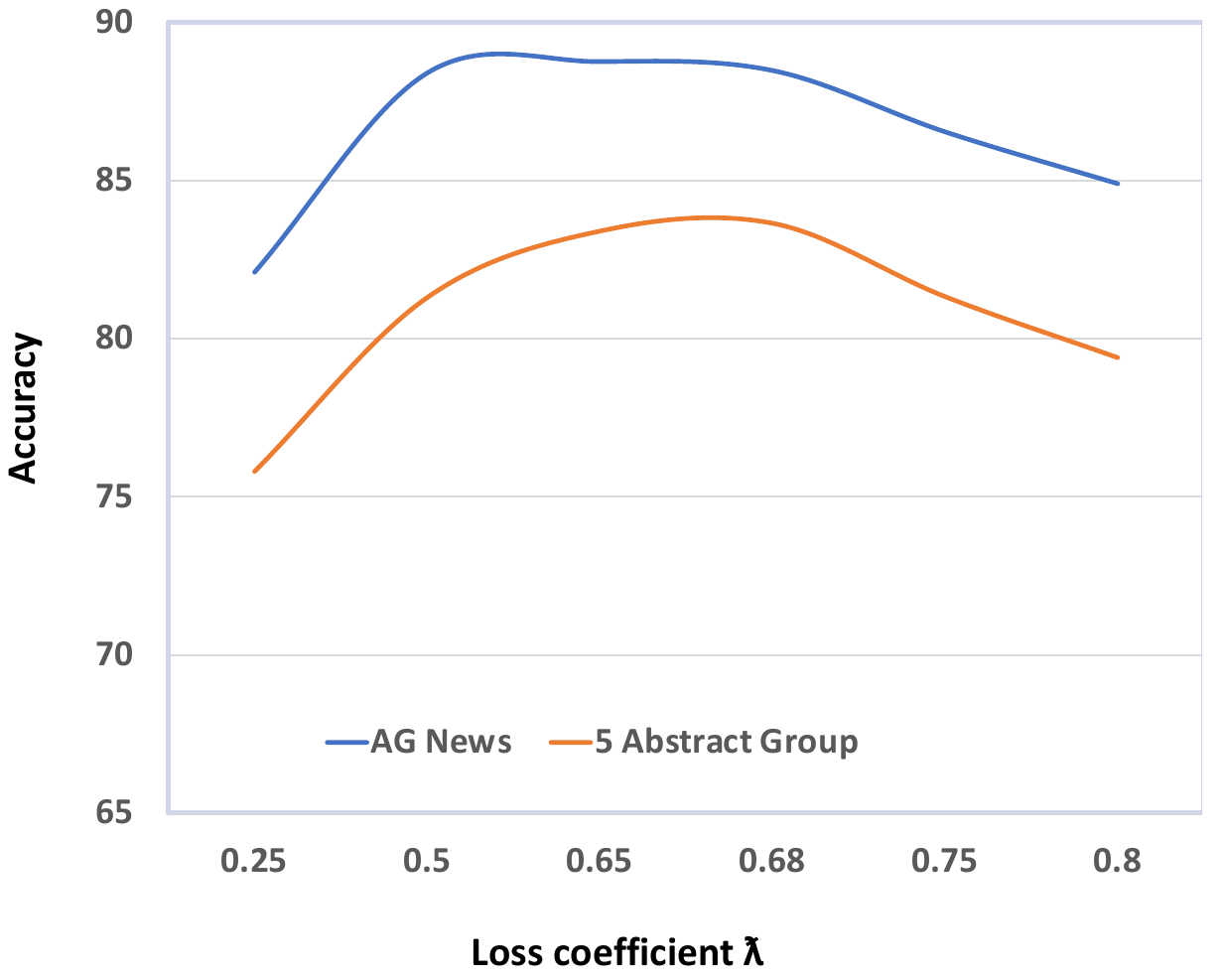}
\caption{Effect of modifying $\lambda$. \label{fig:lambda}}
\end{figure}

\noindent \textbf{Importance of Intermediate Layers:} We conduct a sensitivity analysis by varying $\lambda$ in the loss function, which is associated with the knowledge from intermediate layers. Figure \ref{fig:lambda} presents the effects of different $\lambda$ values on the AG News and 5 Abstracts Group datasets. We find that the model performs best with $\lambda\sim0.65$, indicating the relatively higher importance of intermediate layers for improving performance. This finding aligns with prior studies \cite{liu-etal-2019-linguistic,rogers2021primer}, which have observed that Transformer-based models often encode transferable features in their intermediate layers.

\subsection{Impact of Steerable Data Generation}
\label{sec:app_llm}
\begin{figure}[t!]
\centering
\includegraphics[width=0.5\textwidth]{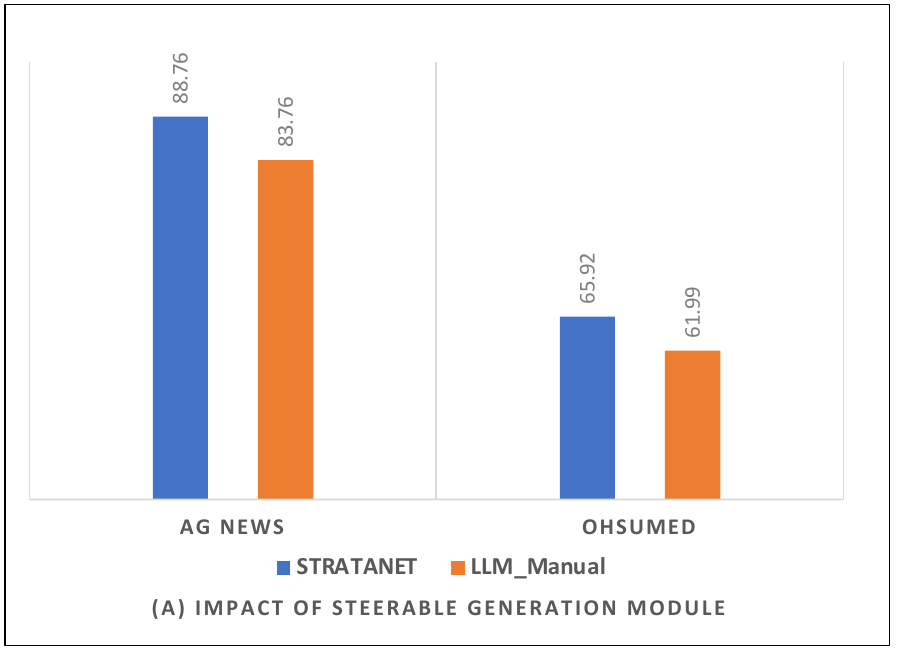}
\caption{Effect of Steerable Data Generation. Llama-2 with manually designed prompts doesn't outperform our generation module. \label{fig:charts_llm}}
\end{figure}

We evaluate the impact of the Steerable Data generation module through $\textsc{Llm}_{Manual}$, involving manual prompting of an LLM like Llama-2 \cite{touvron2023llama} using task-specific prompts and employing diversification techniques (DTs) like sampling variations and temperature adjustments as described in \cite{chung2023increasing}. Figure \ref{fig:charts_llm} shows no significant performance improvement with a more potent Llama-2 model. While relying solely on manual prompting may lack dataset diversity, diversification techniques enhance performance but might introduce irrelevant tokens, impacting overall generation accuracy. Details of the manually designed prompts are given below.

\subsubsection{Manually-designed Prompts}
Table \ref{tab:manual_prompts} show samples of the manually designed prompts to the Llama-2 model. 

\begin{table}[]
\begin{tabular}{@{}l|l@{}}
\toprule
\multicolumn{1}{c|}{\textbf{Datasets}} & \multicolumn{1}{c}{\textbf{Manual Prompts}}                           \\ \midrule
AG News                                & Generate a {[}Category{]} news \textless{}article/story\textgreater{} \\
DBPedia                                & Generate a document about {[}Category{]}                              \\
IMDb                                   & Generate a {[}Category{]} movie review                                \\
SST-2                                  & Generate a {[}Category{]} sentence                                    \\
OhSumed                                & Generate an abstract about {[}Category{]}                             \\ \bottomrule
\end{tabular}
\caption{Samples of dataset-specific manually designed prompts provided as input to the Llama-2 (llama-2-70b-chat) model. \label{tab:manual_prompts}}
\end{table}

\subsubsection{Generation Parameters}
For diversification, we use different temperature setting while we sample tokens. We used five temperature values $\rho \in \{0.3,0.5,0.7,0.9,1.3\}$. Furthermore, we also experimented with different sampling techniques. For nucleus sampling, we varied the top-$p$ between $\{0.65,0.95\}$. For top-k sampling, we chose $k \in  \{10,25,35,50,75\}$.




\end{document}